# MCFNet: Multi-scale Covariance Feature Fusion Network for Real-time Semantic Segmentation

Xiaojie Fang [1]
Mechanical Engineering School, Southwest Jiaotong University, Chengdu 610031, china;
e-mail: fxjswjtu@my.swjtu.edu.cn

Xingguo Song [2*]
Mechanical Engineering School, Southwest Jiaotong University, Chengdu 610031, china;
*e-mail: xg.song@hotmail.com

Xiangyin Meng [3]
Mechanical Engineering School, Southwest Jiaotong University, Chengdu 610031, china;
e-mail: xymeng@swjtu.cn

Xu Fang [4]
Mechanical Engineering School, Southwest Jiaotong University, Chengdu 610031, china;
e-mail: 2314738909@qq.com

Sheng Jin [5]
Mechanical Engineering School, Southwest Jiaotong University, Chengdu 610031, china;
e-mail: 2468857535@qq.com

*Abstract*—The low-level spatial detail information and high-level semantic abstract information are both essential to the semantic segmentation task. The features extracted by the deep network can obtain rich semantic information, while a lot of spatial information is lost. However, how to recover spatial detail information effectively and fuse it with high-level semantics has not been well addressed so far. In this paper, we propose a new architecture based on Bilateral Segmentation Network (BiseNet) called Multi-scale Covariance Feature Fusion Network (MCFNet). Specifically, this network introduces a new feature refinement module and a new feature fusion module. Furthermore, a gating unit named L-Gate is proposed to filter out invalid information and fuse multi-scale features. We evaluate our proposed model on Cityscapes, CamVid datasets and compare it with the state-of-the-art methods. Extensive experiments show that our method achieves competitive success. On Cityscapes, we achieve 75.5% mIOU with a speed of 151.3 FPS.

I. INTRODUCTION

Semantic segmentation has been an important part of scene understanding for a long time. The segmentation process can be regarded as a classification task, which classifies each pixel in an image and assigns a corresponding label to each pixel.

Since AlexNet [1] won the ImageNet classification competition, more and more deep learning methods, especially convolutional neural networks (CNNs) such as VGG [2], GoogLeNet [3], ResNet [4] and Xception [5], have achieved excellent results on classification tasks. However, these methods seem to perform less well when faced with intensive classification tasks such as semantic segmentation. The reason for that is still the specificity of the semantic segmentation task, it requires not only low-level spatial detail information that helps generate clear and detailed boundaries but also high-level semantic abstract information that helps classify and identify. Although these methods are able to extract high-level semantic abstract information, they do not perform well in the segmentation domain due to the continuous convolution, pooling, and downsampling that make the spatial details continuously lost, and also do not make good use of these low-level spatial details. Until now, how to effectively extract, recover and fuse spatial detail information is still an important problem in semantic segmentation.

To solve the above-mentioned problems, most of the current approaches are based on the following three methods.

i) Using spatial branch. As the name implies, spatial branch uses a separate small branch to preserve spatial detail information and encode spatial information. The main representative network is BiSeNet [6] The spatial branch in BiSeNet consists of three layers, each of which contains a stride=2 convolution, batch normalization [7], and ReLu [8] activation function. This branch can encode rich spatial detail information by performing three successive downsampling of the original input image.

ii) Using the skip connection. Skip connection means connecting the features of different layers so that the features of different layers can be integrated with each other and also be complemented by each other. Some of the classical networks that use skip connections are DenseNet [9] and U-Net [10], etc. DenseNet is mainly inspired by ResNet and its idea is also relatively simple, the input of each layer includes information from all previous layers and forms a richer description and discrimination of features by combining features from previous layers.

iii) Using deconvolution. Deconvolution, also known as transposed convolution, can be seen as the inverse process of regular convolution. It is often used in the context of semantic segmentation to recover spatial detail information. However, deconvolution can only recover spatial detail information to a certain extent, but not sufficiently. Therefore, deconvolution can only be regarded as a mitigation tool and not an effective solution. In [11], the authors proposed a deep deconvolution network, consisting of deconvolution, upsampling-bilinear interpolation, and the ReLu activation function, which can obtain more accurate segmentation results. Fig. 1 summarizes the principles of the three methods.

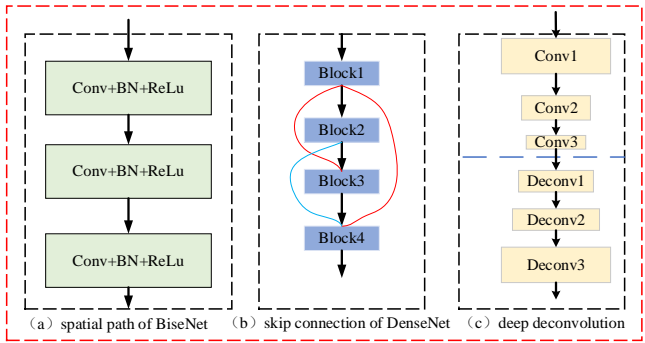

Fig. 1. Demonstration of the three methods.

For the above three methods, we believe that both low-level spatial detail information and high-level semantic abstract information are equally important in the segmentation process, with low-level spatial detail information facilitating the generation of clear and detailed boundaries and high-level semantic abstract information facilitating the classification of categories of images. How to effectively fuse low-level spatial detail information and high-level semantic abstract information is a concern. For this reason, we propose a covariance feature fusion module and a covariance feature refinement module. Meanwhile, we are concerned that the information fusion contains a large amount of invalid information, for this reason, we borrow the idea of gating in LSTM [12][13] networks and propose a gating unit named L-Gate that can filter invalid information effectively while guaranteeing the passage of valid information.

Our main contributions are summarized as follows：

1. We propose a covariance feature fusion module (CFFM), which can learn high and low-level features and to generate weight fusion parameters adaptively, can also achieve the fusion of high-level semantic information and low-level spatial information effectively. CFFM is flexible, which can be used in any network to fuse multi-layer information, and can be cascaded in various ways.

2. We propose a covariance feature refinement module (CFRM), which can refine the features extracted from the network to improve the segmentation accuracy of the network with a negligible cost.

3. We propose L-Gate, a gating unit that can effectively filter invalid information while guaranteeing the passage of valid information. It can filter out invalid information before performing feature fusion and ensure the efficient fusion of valid information to achieve high-resolution prediction.

4. We implement extensive experiments on two datasets, Cityscapes [14] and CamVid [15], to validate the effectiveness of our method.

II. RELATED WORK

*A. Context modeling*

Although convolutional neural networks (CNNs) have achieved impressive performance in extracting image features and also advance the development of semantic segmentation. It is worth noting that semantic segmentation has a greater requirement for receptive field. Therefore, contextual modeling is important for semantic segmentation. Most of the existing methods use atrous convolution, spatial pyramids, and dense concatenation to obtain rich contextual information. For example, Deeplab family [16][17] proposed an Atrous Spatial Pyramid Pooling module (ASPP) based on atrous convolution to capture rich contextual information by using atrous convolution with different dilation rates. The DenseNet proposed by Huang et al [9] uses the idea of dense connectivity at its backbone network, which can extract rich contextual information. The GFFNet proposed by Li et al [18] also adopts the idea of dense connectivity in the network, but unlike DenseNet which adopts dense connectivity in its backbone. GFFNet adopts dense connectivity in the decoding process, which can effectively fuse the contextual information extracted from the backbone. The PSPNet proposed by Zhao et al [19] uses a spatial pyramid pooling module to achieve contextual information aggregation on multiple scales. Elhassan et al proposed S²-FPN [20], it uses a spatial pyramid module for contextual information extraction. At the same time, in order to improve the effectiveness of fusion, spatial attention fusion module is also proposed, which can realize the aggregation of multi-scale context information.

*B. Gating*

Gating units are very common in recurrent neural networks, and the propagation of information can be controlled by them. For example, LSTM has forgetting gates, input gates, and output gates. Forgetting gates can decide what information is discarded from the cell state, input gates can decide what new information is stored into the cell state, and output gates can decide what information can be output. By using different gates, the problem of long-term dependence can be handled. The GSCNN proposed by Takikawa et al [21] also uses gating units, which are used to learn the exact boundary information.

In this work, we are inspired by the LSTM network, and using the feature that the gating unit can control the information propagation, we propose L-Gate, which can realize the filtering of a large amount of invalid information, thus improving the effect of model information fusion.

*C. Real-time semantic segmentation*

Real-time semantic segmentation is a method that places high demands on the speed of segmentation and generates high-quality predictions. The real-time semantic segmentation network CGNet [22] is a lightweight context-guided network that proposes CG block for learning joint features of local features and surrounding contexts. In addition, CG block can capture contextual information at all stages efficiently. ContextNet [23] is similar to the baseline network BiseNet used in this paper, which also employs two branches. The low-resolution branch captures contextual information and the high-resolution branch captures spatial detail information. The authors in [24] concluded that the backbone network for image classification is not particularly suitable for the semantic segmentation task, so they carefully designed the backbone network STDCNet (Short-Term Dense Concatenate Network). To extract rich spatial detail information, the authors removed the spatial detail branch and replaced it with a detail-guided module that allows the network to learn how to extract to spatial detail information effectively.

## III. SEGMENTATION NETWORK

### A. Spatial path

Spatial detail information is very important in semantic segmentation, spatial detail branch is mainly responsible for encoding spatial detail information. The rich spatial detail information is the basis of segmentation. Since the constant convolution and pooling will lose the image details and contour information, the spatial detail branch can not be designed too deep. Here, we still design the spatial branch as 3 layers. To extract the spatial information better, we replace the 3×3 convolution kernel in the first layer of convolution with a 7×7 convolution kernel to expand the field-of-view.

### B. Context path

Both low-level spatial detail information and high-level abstract semantic information are necessary for semantic segmentation. The spatial detail branch can encode rich spatial detail information. For encoding high level of abstract semantic information, existing methods use continuous convolutional downsampling and pooling, such as U-Net, DeepLab, etc. As for encoding rich contextual information, here, we use the contextual path network from BiseNet. In this work, the backbone network we use is ResNet18, for obtaining large receptive field and encoding high-level semantic context information, we use a global average pooling operation at the end of the context path, which provides the global maximum receptive field. In the end, when fusing different feature information, we use the CFFM method proposed in this work, which can efficiently and adaptively fuse the information from different layers.

### C. Covariance feature refinement module

Covariance is used in probability statistics to measure the overall error of two variables. The covariance between two variables is positive if one is greater than its own expected value and the other is also greater than its own expected value. The covariance between two variables is negative if they have opposite trends, i.e., one is greater than its own expected value and the other is less than its own expected value. In computer vision, different network layers output different feature layers, but there is some connection between different featured points. In order to allow efficient fusion of different feature layers and thus improve the accuracy, this work proposes a covariance feature fusion module (CFFM) and a covariance feature refinement module (CFRM) to generate feature integration vectors by calculating the covariance between high and low feature layers and then adding them to two feature layers, which can adaptively guide the efficient fusion between different feature layers. The covariance is calculated as Eq. (1), X and Y are the feature point vectors of different layer features, respectively. $\overline{X}$ and $\overline{Y}$ are their corresponding mean values, n is the number of feature points, $\mathrm{cov}(X,Y)$ is the covariance coefficient.

$$\mathrm{cov}(X,Y) = \frac{\sum_{i=1}^{n}(X_i - \overline{X})(Y_i - \overline{Y})}{n-1} \quad (1)$$

To improve the segmentation accuracy of the segmentation network, we propose a CFFM module, which can achieve efficient fusion of information from high and low feature layers in segmentation adaptively. It can learn the features of different feature layers, generate feature fusion vectors, and guide the model to complete convergence.

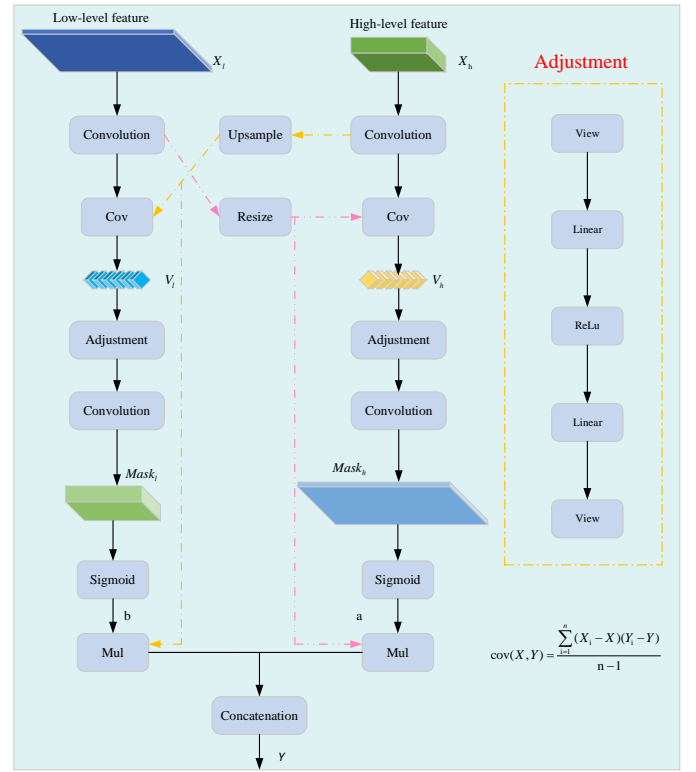

Fig. 2. Covariance feature fusion module.

The structure of the CFFM module is shown in Fig. 2. From this we can see that in order to facilitate the fusion of high-level and low-level information, we adopt two branches. In the right branch, we calculate the covariance between the high-level feature map and the sampled low-level feature map, generating the covariance vector.

After that, we send the vector to the adjustment module, where the structure of the adjustment module is shown in Fig. 2. The output of the adjustment module is convolved to generate the feature mask, which is passed through the sigmoid function to generate the a-vector, the output of the right branch is obtained by multiplying the a-vector with the low-level feature map. The output of the left branch is similar to the right branch.

The working process of the right branch can be described in Eq. (2).

$$V_h = \mathrm{CoV}\{\mathrm{Conv}_{[1\times 1, C_h]}(X_h, \theta_h), R[\mathrm{Conv}_{[1\times 1, c_l]}(X_l, \theta_l)]\}$$
$$a = \mathrm{Sig}(\mathrm{Conv}_{[3\times 3]}[\mathrm{Adj}(V_h)]) \quad (2)$$

The left branch works in a similar way to the right branch and can be described by the following equation.

$$V_l = \mathrm{CoV}\{\mathrm{Up}[\mathrm{Conv}_{[1\times 1, c_h]}(X_h, \theta_h)], \mathrm{Conv}_{[1\times 1, c_l]}(X_l, \theta_l)\}$$
$$b = \mathrm{Sig}(\mathrm{Conv}_{[3\times 3]} \mathrm{Adj}(V_l)) \quad (3)$$

The output of the CFFM module can be summarized in Eq. (4).

$$Y = \mathrm{Cat}(a \times X_h, b \times X_l) \quad (4)$$

In Eqs. (2), (3), and (4), $X_l$, $X_h$ denotes the high-level semantic information and the low-level spatial information, respectively, Y denotes the output of the module, $\mathrm{Conv}_{[1\times 1, C_h]}$ denotes the convolution kernel in the convolution operation is $1 \times 1$, and the input $C_h$, $\theta_h$ denotes the parameters learned in this convolution process, $\mathrm{Adj}(\bullet)$ denotes the operation by the Adjustment module, $\mathrm{Up}(\bullet)$ and $\mathrm{Sig}(\bullet)$ denote the upsampling operation and the sigmoid operation, respectively.

### D. Covariance feature refinement module

To refine the features extracted by the model, we propose a covariance feature refinement module inspired by the attention mechanism [25][26], which can effectively refine the features extracted by different layers. By taking the method of covariance coefficients and calculating the covariance attention vector to guide the model to accomplish feature learning. This module (CFRM) can refine the output of the feature at each stage of the model, without any upsampling operation, can easily accomplish the feature refinement function. The computational cost of this module is negligible but brings an increase in the accuracy. The subsequent experimental phase will use ablation experiments to demonstrate the effectiveness of the module proposed in the paper.

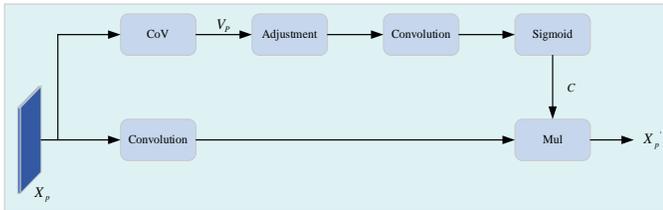

Fig. 3. Covariance feature refinement module.

The overall structure of the CFRM module is shown in Fig. 3. The working process of the entire covariance feature refinement module can be briefly summarized as follows. First, the input of the module $X_p$ and itself calculate the covariance feature vector $V_p$ and then $V_p$ passes through the adjustment module, the convolution, and the sigmoid function to generate the vector C. The output of CFRM module is obtained by multiplying C with $X_p$.

The workflow of this module can be described in Eq. (5), $X_p$ denotes the input feature layer of this module, $V_p$ denotes the covariance feature vector $X_p'$ computed by this module, $X_p'$ denotes the output of this module.

$$V_p = \mathrm{CoV}(X_p, X_P)$$
$$c = \mathrm{Sig}(\mathrm{Conv}_{[3\times 3]}(\mathrm{Adj}(V_p))) \quad (5)$$
$$X_p' = c \times \mathrm{Conv}_{[3\times 3, C_p]}(X_p)$$

### E. L-Gate

To improve the fusion efficiency and filter out a large amount of invalid information. We are inspired by the Long Short-Term Memory Network (LSTM) and propose a gating unit named L-Gate. The structure of L-Gate is shown in Fig. 4, the information from each way will pass through the gating unit, thus filtering out the invalid information. Finally, we fuse the information of each way to get our output.

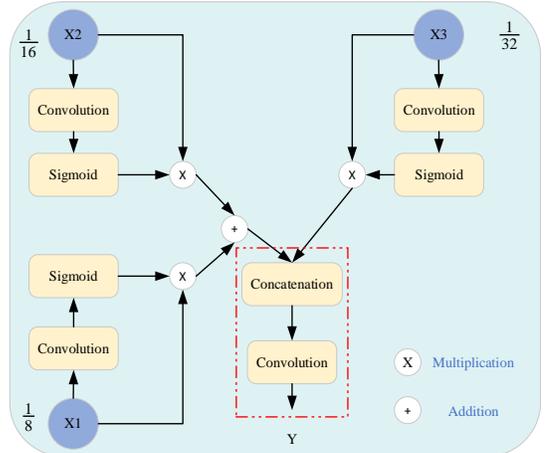

Fig. 4. L-Gate structure.

The output of L-Gate can be described in Eq. (6), $Y$ denotes the output of L-Gate, $x_i$ denotes the input of the ith forgetting gate, $\mathrm{Cat}(\bullet)$ denotes the concatenation operation, $\mathrm{mul}(\bullet)$ denotes a multiplication operation and $f_i(\bullet)$ denotes the operation of the ith forgetting gate.

$$Y = \mathrm{Conv}_{[3\times 3]}\{\mathrm{Cat}[f_1(X_1) + f_2(X_2), f_3(X_3)]\}$$
$$f_i(x) = \mathrm{mul}(x_i, \mathrm{Sig}(\mathrm{Conv}_{[3\times 3]}(x_i))) \quad (6)$$

### F. Network Architecture

The general structure of our network is shown in Fig. 5, it can be seen that our network structure is also composed of two branches, a spatial detail branch, and a context branch. Unlike [6], we replace the convolution kernel of the first layer of convolution into 7×7 in the spatial branch. In addition, in the context branch, we utilize the 1/8 feature layer as well. Then, we use 1/8 ,1/16 and 1/32 feature maps as the input of our L-Gate proposed in this work. We fuse the output of L-Gate with the output of spatial detail branch in the covariance feature fusion module. Finally, we add the refinement module of this work following the feature fusion module proposed in BiseNet, it can refine the features and obtain high-accuracy predictions.

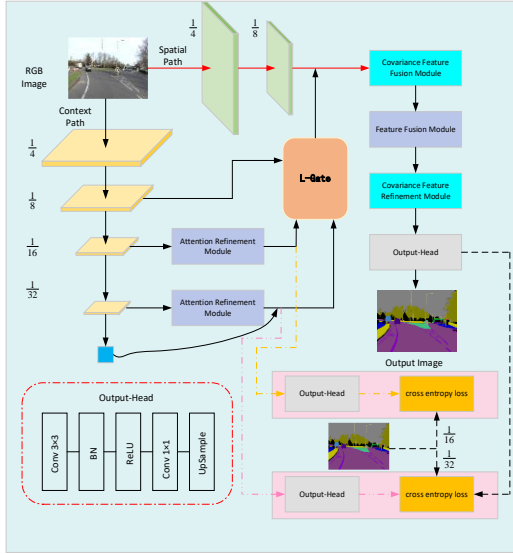

Fig. 5. Structure of MCFNet.

## IV. EXPERIMENTAL RESULTS

We will validate our proposed modules on two urban street scenes data sets, Cityscapes and CamVid. In this section, we will first introduce the two datasets and the evaluation metrics, separately. Then, we will present the details of our experiments and conduct a series of ablation experiments to validate the effectiveness of our proposed modules. To be more convincing, we will compare our model with other semantic segmentation models. Finally, we will discuss the impact of our proposed modules.

### A. Datasets and evaluation Metrics

Cityscapes: The Cityscapes dataset has 5000 images of driving scenes in urban environments (2975 for training, 500 for validation, and 1525 for testing), recording street scenes in 50 different cities. It contains 30 categories, 19 of which are used for the semantic segmentation task. With image resolutions up to 2048 × 1024, this dataset is authoritative in semantic segmentation task. For a fair comparison with other algorithms, we only use finely annotated images here.

CamVid: The CamVid dataset is a dataset of urban road scenes where the producers of the dataset collect video from the perspective of driving. It contains 701 annotated images extracted from the video sequences, 367 images are used for training, 101 images for validation, and 233 images for testing. The image resolution is 960×720, and there are 32 semantic categories, 11 of which are used for the semantic segmentation task.

Evaluation Metrics: We will take the standard metric of the mean intersection over union (mIOU) and frames per second (FPS) as our evaluation metrics. The Intersection over Union (IoU) is the ratio of the intersection of the true label and the predicted value for that class to their union, mIOU is defined as the mean IoU for each class in the dataset. The FPS is defined as the number of frames processed by the model per second. mIOU can be expressed in Eq. (7),

$$mIOU = \frac{1}{k+1}\sum_{i=0}^{k}\frac{p_{ii}}{\sum_{j=0}^{k}p_{ij}+\sum_{j=0}^{k}p_{ji}-p_{ii}} \quad (7)$$

where $k$ denotes the image category, $k+1$ is the image category plus the background, $p_{ii}$ is the number of correctly classified pixel category $i$, and $p_{ij}$ is the number of misclassified pixel category $i$ to pixel category $j$.

### B. Implementation protocol

In this section, we will introduce the details of our implementation.

Network: We use ResNet18 to extract contextual information. To extract rich spatial information and expand the perceptual field, we introduce residual connectivity and global average pooling. Finally, to fuse the features of the two branches efficiently, we propose a covariance feature fusion module and a covariance feature refinement module as well as an L-Gate unit to obtain the final prediction results.

Training details: We use mini-batch stochastic gradient descent (SGD) with batch size 4, momentum 0.9, and weight decay 1e−4 in training. We use a "poly" learning rate strategy with a power of 0.9 for each iteration. The initial learning rate is 2.5e-2. In addition, we use online hard example mining (OHEM) [27] and linear warmup training method [28] for the first 3000 iterations on CityScapes. The "poly" learning rate strategy and the linear warmup training method can be illustrated in Eq. (8) and Eq. (9), where lri is the initial learning rate, iter is current iteration, lrmax denotes the maximum learning rate, warmup_factor is the ratio between the maximum learning rate and the initial learning rate, in our training process, we set warmup_factor=0.1.

$$lr = lr_i \times (1-\frac{iter}{max\_iter})^{power} \quad (8)$$

$$lr = lr_{max} \times warmup\_factor^{1-\frac{iter}{warmup\_iters}} \quad (9)$$

Loss: We use the cross-entropy loss function, it can be described in Eq. (10) and Eq. (11), where ti is the true value, pi is the output prediction of our model.

$$loss = -\sum_{i=1}^{n}t_i \log(p_i) \quad (10)$$

$$p_i = \frac{e^{f_i}}{\sum_j e^{f_j}} \quad (11)$$

Data augmentation: We use random horizontal flipping, random cropping, random scaling and color jittering. The scale ranges in [0.5,1.0,1.25,1.5,1.75] and cropped resolution is 1024×512 for training Cityscapes. For CamVid, the scale ranges in [0.5,1.0,1.25,1.5] and cropped resolution is 960×720, In addition, we only use 11 classes and set other classes to the ignore label=255.

*C. Ablation study*

In this section, we will validate the effectiveness of our proposed modules through three ablation experiments for L-Gate, CFFM, and CFRM, respectively. We train our model using the PyTorch 1.11 framework with a NVIDIA RTX 1660Ti GPU under CUDA 11.4 and CUDNN 8.5.

Baseline: We use the BiseNet as our baseline model, the dataset we use in doing the ablation experiments is CityScapes and the backbone network is ResNet18.

Ablation for L-Gate: In Sec. (2), we propose a gating unit L-Gate to improve the fusion efficiency and filter out a large amount of invalid information. We use the lightweight model ResNet18 as our backbone of the context path, meanwhile, we apply our L-Gate before fusing the semantic information and the spatial detail information, as shown in Table 1, we increase the mIOU of the baseline model from 73.7% to 74.2% with only 0.02M extra parameters.

Ablation for CFFM Module: In order to enhance the fusion efficiency of semantic and spatial information, we propose CFFM in this work, it can be observed that the mIOU increases from 74.2% to 74.6% when adding CFFM in our ablation experiments.

Ablation for CFRM Module: In order to further improve the performance of our model, we introduce CFRM based on CFFM, as shown in Table 1, it does not bring additional parameters but increases the mIOU to 75.5%.

TABLE I. ACCURACY AND PARAMETERS ANALYSIS OF OUR BASELINE MODEL AND OUR ABLATION EXPERIMENTS.

| Method | Parameters | Flops | mIoU | Δa(%) |
|---|---|---|---|---|
| Baseline | 13.43M | 30.86G | 73.7 | 0.0 |
| Baseline+ L-Gate | 13.45M | 31.07G | 74.2 | 0.5↑ |
| Baseline+ L-Gate+CFFM | 13.70M | 32.55G | 74.6 | 0.9↑ |
| Baseline+ All modules | 13.70M | 34.34G | 75.5 | 1.8↑ |

*D. Compare with state-of-the-arts*

In this section, we compare our method with other existing semantic segmentation methods on two datasets, Cityscapes, and CamVid, respectively. The measurement of inference time is executed on NVIDIA GeForce GTX 3060 with CUDA 11.4, CUDNN 8.5 and TensorRT 8.2.5. In addition, we use FP32 data precision to calculate the inference time here.

Table 2 shows the result of our method and other 15 models on CityScapes. As can be seen from Table 2, we present the model name, input resolution, backbone, mIOU and FPS of various approaches, our proposed model MCFNet can achieve 151.3 FPS and a 75.5% mIOU on CityScapes validation set. BiseNetV1 with the resolution 512×1024 is our baseline, Table 2 reports that BiSeNetV1 with the resolution 768×1536 can achieve 65.5 FPS and a 74.8% mIOU, MCFNet is 0.9% more than BiSeNetV1 on mIOU, while having 1.3 times more inference speed. When compared to two recent methods, BiSeNetV3 and CIDNet, MCFNet is 2.9% more than BiSeNetV3-50 and 0.5% more than CIDNet1-50 on mIOU.

Table 3 displays the performance on CamVid data set, for better tensor size compatibility with our implementation, we resize input images to 720×960 during training and evaluating phase. Similar to Table 2, we also present the model name, input resolution, backbone, mIOU and FPS of various approaches. Our proposed MCFNet can achieve 145.5 FPS and a 76.0% mIOU on CamVid. Compared to BiSeNetV1, our model achieves a 10.6% higher mIOU while being 0.25 times faster on FPS. At the same time, we compare MCFNet with CIDNet and BFMNet, it can be seen that MCFNet is 6.6% more than CIDNet and 2.2% more than BFMNet1 on mIOU, meanwhile, MCFNet is faster than CIDNet and BFMNet on CamVid.

TABLE II. ACCURACY COMPARISON OF OUR MODEL AGAINST OTHER STATE-OF-THE-ART METHODS ON CITYSCAPES.

| Model | Resolution | Backbone | mIoU (%) val | mIoU (%) test | FPS |
|---|---|---|---|---|---|
| BiSeNetV1 [6] | 768×1536 | Xception39 | 69.0 | 68.4 | 105.8 |
| BiSeNetV1 [6] | 768×1536 | ResNet18 | 74.8 | 74.7 | 65.5 |
| STDC1-Seg50 [24] | 512×1024 | STDC1 | 72.2 | 71.9 | 250.4 |
| STDC2-Seg50 [24] | 512×1024 | STDC2 | 74.2 | 75.3 | 188.6 |
| BiSeNetV2 [29] | 512×1024 | No | 73.4 | 72.6 | 156 |
| BiSeNetV2-L [29] | 512×1024 | No | 75.8 | 75.3 | 47.3 |
| DFANet A' [30] | 512×1024 | Xception A | - | 70.3 | 160 |
| DFANet A [30] | 1024×2048 | Xception A | - | 71.3 | 100 |
| FasterSeg [31] | 1024×2048 | No | 73.1 | 71.5 | 163.9 |
| CAS [32] | 768×1536 | No | - | 70.5 | 108.0 |
| GAS [33] | 769×1537 | No | - | 71.8 | 108.4 |
| BiSeNetV3-50 [34] | 512×1024 | STDC1 | 73.4 | 73.5 | 244.3 |
| BiSeNetV3-50 [34] | 512×1024 | STDC2 | 74.6 | 74.5 | 180.3 |
| CIDNet1-50 [35] | 512×1024 | No | 75.1 | 73.5 | 164.1 |
| CIDNet2-50 [35] | 512×1024 | No | 75.3 | 74.4 | 139.0 |
| MCFNet | 512×1024 | ResNet18 | 75.5 | 75.3 | 151.3 |

TABLE III. ACCURACY COMPARISON OF OUR MODEL AGAINST OTHER STATE-OF-THE-ART METHODS ON CAMVID. NO INDICATES THE METHOD DO NOT HAVE ITS BACKBONE OR THE AUTHOR DIDN'T NAME THEIR BACKBONE.

| Model | Resolution | Backbone | mIoU(%) | FPS |
|---|---|---|---|---|
| BiSeNetV1 [6] | 720 × 960 | Xception39 | 65.6 | 175 |
| BiSeNetV1 [6] | 720 × 960 | ResNet18 | 68.7 | 116.3 |
| STDC1-Seg [24] | 720 × 960 | STDC1 | 73.0 | 197.6 |
| STDC2-Seg [24] | 720 × 960 | STDC2 | 73.9 | 152.2 |
| BiSeNetV2 [29] | 720 × 960 | No | 72.4 | 124.5 |
| BiSeNetV2-L [29] | 720 × 960 | No | 73.2 | 32.7 |
| CAS [32] | 720 × 960 | No | 71.2 | 169.0 |
| GAS [33] | 720 × 960 | No | 72.8 | 153.1 |
| BiSeNetV3 [34] | 720 × 960 | STDC1 | 75.1 | 198.4 |
| CIDNet1 [35] | 720 × 960 | No | 71.3 | 130.7 |
| CIDNet2 [35] | 720 × 960 | No | 71.5 | 109.6 |
| BFMNet1 [36] | 720 × 960 | MobileNetV3 | 74.4 | 118.6 |
| BFMNet2 [36] | 720 × 960 | ResNet18 | 75.6 | 95.8 |
| MCFNet | 720 × 960 | ResNet18 | 76.0 | 145.5 |

In order to demonstrate the superiority of our proposed model MCFNet, we give the visualization results of BiseNetV1,

BiseNetV2 and our model on CityScapes validation and test sets. As shown in Fig. 6 and Fig. 7, each column from left to right is input images, the prediction results of BiseNetV1, BiseNetV2 and our MCFNet. For a better comparison, we use yellow rectangular boxes to frame the same object in the picture, it can be seen that the prediction results of MCFNet are superior to BiseNetV1 and BiseNetV2.

Based on Table 2 and Table 3, we plot Fig. 8 and Fig .9, respectively, the horizontal axis is FPS and the vertical axis is mIOU. In the two figures, we use different colors of lines and markers to represent different models, with our model being represented by a red asterisk. We can observe that our MCFNet achieves the trade-off between speed and performance on the two datasets.

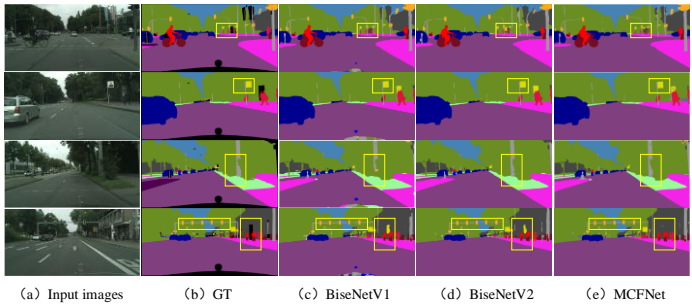

(a) Input images  (b) GT  (c) BiseNetV1  (d) BiseNetV2  (e) MCFNet

Fig. 6. The visual comparison of our model (MCFNet) with BiseNetV1 and BiSeNetV2 on Cityscapes validation set.

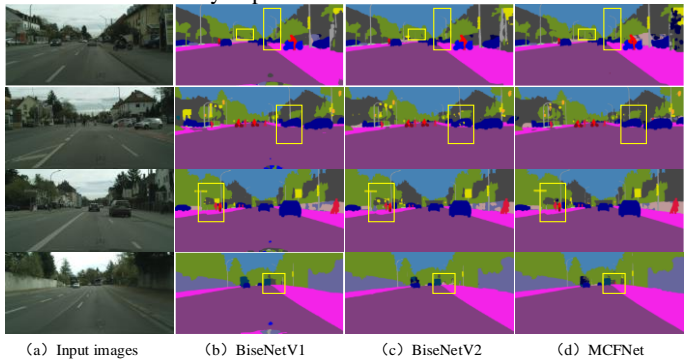

(a) Input images  (b) BiseNetV1  (c) BiseNetV2  (d) MCFNet

Fig. 7. The visual comparison of our model (MCFNet) with BiseNetV1 and BiSeNetV2 on Cityscapes test set.

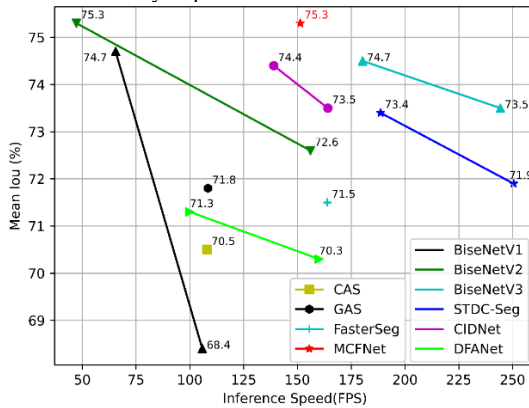

Fig. 8. FPS vs mIOU on Cityscapes dataset. Our model is in red.

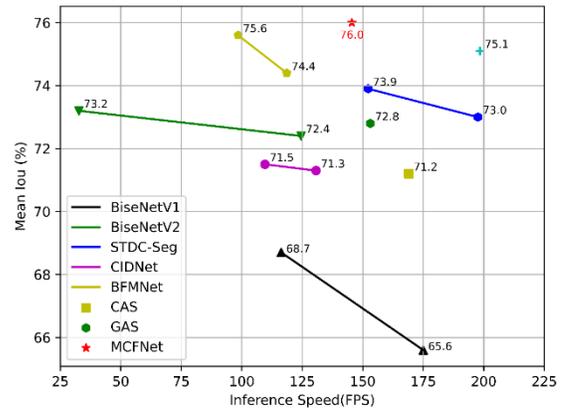

Fig. 9. FPS vs mIOU on CamVid dataset. Our model is in red.

## V. Conclusions

To fuse the low-level spatial detail information with the high-level semantic abstract information efficiently, we propose a covariance feature fusion module (CFFM) in this paper. At the same time, to refine the features extracted from the model, we also propose a covariance feature refinement module (CFRM) in this paper, which can adaptively find the association between different feature layers to achieve efficient fusion between different features. In addition, we note that the perceptual field plays a crucial role in the segmentation process, we use the global average pooling operation at the end of the context branch to obtain the maximum perceptual field in the whole model. Finally, we find that there is a lot of invalid information in the fusion process, we propose a gating unit named L-Gate to improve the efficiency and accuracy of the fusion. We test our model on Cityscapes, CamVid datasets, our model achieve competitive results. In the future, our work will be extended to the following aspects: (i) We will validate the effectiveness of our proposed module on other computer vision tasks, e.g., object detection, and image classification. (ii) We will continue to explore more effective methods for spatial detail information extraction on semantic segmentation tasks. (iii) We will continue to investigate more efficient feature fusion methods. (iv) We observe that the ResNet network is specifically designed for image classification tasks, too many image categories are easy to cause the structural redundancy and channel redundancy in semantic segmentation. We will investigate the backbone network suitable for semantic segmentation.

ACKNOWLEDGMENT

This work is supported by the Fundamental Research Funds for the Central Universities (No. 2682022KJ015), State Key Laboratory of Robotics and Systems (HIT) (SKLRS-2020-KF-13).